\documentclass{article}

\usepackage{microtype}
\usepackage{graphicx}
\usepackage{subcaption}
\usepackage{booktabs}
\usepackage{hyperref}
\usepackage{xurl}

\usepackage[preprint]{icml2026}
\makeatletter
\renewcommand{\Notice@String}{}
\makeatother
\usepackage{amsmath}
\usepackage{amssymb}
\usepackage{mathtools}
\usepackage{array}
\usepackage{newfloat}
\usepackage{listings}
\DeclareCaptionStyle{ruled}{labelfont=normalfont,labelsep=colon,strut=off}
\floatstyle{ruled}
\newfloat{listing}{tb}{lst}{}
\floatname{listing}{Listing}
\icmltitlerunning{Library Reachability in LSR-Synth}

\begin{document}

\twocolumn[
\icmltitle{\texorpdfstring{Library Reachability in LSR-Synth:
How Anti-Memorization Design\\
Changes the Measurement of Symbolic Discovery}
{Library Reachability in LSR-Synth: How Anti-Memorization Design
Changes the Measurement of Symbolic Discovery}}

\icmlsetsymbol{equal}{*}
\begin{icmlauthorlist}
  \icmlauthor{Zhan'ao Yao}{equal,sic,cmsoe}
  \icmlauthor{Liang Yin}{equal,sic,soais}
  \icmlauthor{Zhihao Gao}{equal,sic,cmsoe}
  \icmlauthor{Boxuan Zhang}{equal,casia,ucasaai}
  \icmlauthor{Xiaoyu Wu}{wenge}
  \icmlauthor{Linjing Li}{casia,ucasaai}
  \icmlauthor{Rongyan Wang}{sic,cmsoe}
  \icmlauthor{Tingwei Chen}{lnu}
  \icmlauthor{Youwei Wang}{sic,cmsoe}
  \icmlauthor{Xiaolin Zhao}{sic,cmsoe}
  \icmlauthor{Jiahui Shi}{casia,ucasaai}
  \icmlauthor{Jianjun Liu}{sic,cmsoe}
\end{icmlauthorlist}

\icmlaffiliation{sic}{State Key Laboratory of High Performance Ceramics,
Shanghai Institute of Ceramics, Chinese Academy of Sciences,
1295 Dingxi Road, Shanghai 200050, China}
\icmlaffiliation{cmsoe}{Center of Materials Science and Optoelectronics Engineering,
University of Chinese Academy of Sciences, Beijing 100049, China}
\icmlaffiliation{soais}{School of Advanced Interdisciplinary Sciences,
University of Chinese Academy of Sciences, Beijing, China}
\icmlaffiliation{casia}{State Key Laboratory of Multimodal Artificial Intelligence Systems,
Institute of Automation, Chinese Academy of Sciences, Beijing 100190, China}
\icmlaffiliation{ucasaai}{School of Artificial Intelligence,
University of Chinese Academy of Sciences, Beijing 100049, China}
\icmlaffiliation{wenge}{Beijing Wenge Technology Co., Ltd., Beijing, China}
\icmlaffiliation{lnu}{Faculty of Information, Liaoning University, Shenyang, China}

\icmlcorrespondingauthor{Tingwei Chen}{twchen@lnu.edu.cn}
\icmlcorrespondingauthor{Youwei Wang}{ywwang@mail.sic.ac.cn}
\icmlcorrespondingauthor{Xiaolin Zhao}{zhaoxiaolin@mail.sic.ac.cn}
\icmlcorrespondingauthor{Jiahui Shi}{jiahui.shi@ia.ac.cn}
\icmlcorrespondingauthor{Jianjun Liu}{jliu@mail.sic.ac.cn}

\icmlkeywords{symbolic regression, scientific discovery, large language models,
benchmark evaluation, memorization}

\vskip 0.3in
]

\printAffiliationsAndNotice{\icmlEqualContribution}

\begin{abstract}
Existing benchmarks for scientific equation discovery are largely composed of well-known equations available in the public domain, making it difficult to determine whether a model is discovering laws from data or merely recalling answers from its training corpus. LSR-Synth mitigates this problem by introducing novel synthetic terms into established scientific mechanisms and filtering the resulting tasks for novelty, solvability, and scientific plausibility. This paper examines a narrower measurement question: can these tasks further distinguish scientific priors supplied by language models from conventional operator search that does not access task semantics? We construct a semantics-free baseline using a fixed vocabulary with publicly documented provenance, and assess the role of candidate coverage through semantic blinding, library weakening, and matched operator-family knockouts. Under the current task snapshot, search budget, and scoring protocol, the fixed vocabulary already covers most tasks, while language-model-generated candidates rarely expand the set of solvable instances. Their marginal contribution becomes substantial only when vocabulary coverage is selectively disrupted. Strict out-of-distribution evaluation lowers the absolute success rates of all methods but does not alter this relationship. These findings neither invalidate LSR-Synth’s controls against memorization of complete formulas nor imply that language-model priors are generally unhelpful. Rather, they support a more limited conclusion: most current tasks remain suitable for evaluating the fitting and recombination of previously unseen expressions, but are insufficient on their own to identify contributions from priors beyond a fixed search space.
\end{abstract}


\section{Introduction}

When large language models are applied to symbolic regression, an immediate evaluation concern is that test equations may already have appeared in the models’ training corpora. Traditional symbolic regression evaluates candidate expressions primarily in terms of numerical error and expression complexity \cite{lacava2021srbench}, whereas language models may also recall formulas directly from variable names, problem descriptions, and prior scientific knowledge. Equations and their accompanying explanations from widely used datasets such as Feynman are extensively represented in publicly available texts \cite{udrescu2020aifeynman,shojaee2025llmsrbench}. Consequently, when a model produces the correct expression, it may either have inferred the underlying relationship from the data or simply recognized the problem and reproduced a known answer; final accuracy alone cannot distinguish between these possibilities \cite{mccoy2019wrongreasons}. LLM-SRBench was developed specifically to address this difficulty by constructing targets that models are unlikely to have encountered verbatim, thereby reducing contamination from formula memorization \cite{shojaee2025llmsrbench}. Within this benchmark, LSR-Synth draws on scientific settings such as reaction kinetics, population growth, oscillatory systems, and constitutive material relationships. It combines established mechanisms with newly generated terms and then filters the resulting tasks for numerical solvability, novelty, and domain plausibility. Variable meanings and scientific context remain available to the language model, allowing it to exploit relevant knowledge, but the complete canonical formulas from the literature are no longer supplied as ready-made answers. Compared with the straightforward reuse of classical equations, this design offers a clear advantage: at least in its complete form, the target is not an existing answer, making it possible to evaluate more reliably a method’s ability to fit and compose previously unseen expressions.

Novelty of the complete equation, however, does not imply novelty of the search space. A high-order saturation term or exponential decay term that is uncommon in a particular scientific context may still be composed entirely of familiar building blocks, such as powers, rational functions, exponentials, and products of variables. A general-purpose search method may therefore already possess all the components required to recover the target, even without access to variable meanings or domain context. At least three levels must be distinguished: whether the complete expression has appeared before; whether the operators and candidate structures composing that expression are already included in the search space; and whether the algorithm can actually find those candidates within the prescribed budget. LSR-Synth primarily controls the first level, whereas whether a language model supplies indispensable scientific priors depends on the latter two. If a task can be solved using a fixed semantics-free library, it remains a valid test of equation fitting and compositional generalization, but it cannot readily distinguish success due to language-model priors from success due to conventional operator search. Only when the fixed library fails and the addition of language-model-generated candidates leads to success is there evidence that the language model has expanded the effective search boundary under the current budget. Such expansion is not limited to inventing new atomic operators: novel nesting patterns, variable transformations, parameterized structures, or more effective compositions may also constitute meaningful contributions. We refer to the property that a target can be recovered by semantics-free search under a prescribed vocabulary and computational budget as *library reachability*. On this basis, we pose a deliberately limited question: under the current evaluation setting, do diagnostic tasks with genuine power to make the above distinction constitute a sufficiently large proportion of the benchmark? We do not use this question to judge LSR-Synth as a whole, nor do we generalize our conclusions to other vocabularies, search algorithms, or symbolic regression problems.

To answer this question, our experiments hold the training data, fitting procedure, selection rules, and search budget fixed while varying only the source of candidates. BANK uses a fixed vocabulary with publicly documented provenance and does not access task semantics; LLM uses candidates proposed by a language model; and UNION searches over the union of the two candidate sets. The comparison between LLM and BANK measures the respective utility of the two candidate sources, whereas the comparison between UNION and BANK tests whether the language model adds candidates beyond those available in the fixed library. Semantic blinding is used to assess the contribution of domain information. A second publicly sourced vocabulary, a polynomial-only library, and matched operator knockouts are used to determine whether the conclusions change with candidate coverage. Evaluation includes in-distribution numerical accuracy, strict joint recovery across in-distribution and out-of-distribution data, and symbolic accuracy. The results show that the semantics-free fixed library already covers most tasks. UNION produces only minor changes when the full library is used, but yields substantial improvements with the polynomial-only library. We therefore make only the following claim: the current aggregate benchmark score is insufficient, on its own, to demonstrate that a language model contributes scientific knowledge beyond what is already encoded in a fixed library.

\section{Related Work}

Symbolic regression benchmarks traditionally compare methods through numerical accuracy, expression complexity, and recovery of known target equations \cite{lacava2021srbench,udrescu2020aifeynman}. This protocol becomes ambiguous for language models because canonical equations and their scientific descriptions may occur in pretraining corpora. LLM-SRBench addresses complete-formula memorization by constructing targets that are unlikely to have appeared verbatim and by retaining scientific context as an input to the model \cite{shojaee2025llmsrbench}. Our analysis addresses a complementary question. Rather than asking only whether the complete target is novel, we test whether its component structures remain reachable by a documented, semantics-free candidate library under the same fitting budget.

LLM-assisted symbolic regression systems also intervene at different stages of discovery. LLM-SR uses language models to write and refine search programs \cite{shojaee2025llmsr}; LaSR supplies a learned concept library to a symbolic search procedure \cite{grayeli2024lasr}; and SR-Scientist combines proposal, experimentation, and evaluation in an agentic workflow \cite{xia2026srscientist}. More recent systems use data-derived structural reasoning, executable priors, retrieval, or guided combinatorial search. Table~\ref{tab:method_map} organizes representative examples by the information used, the way candidates are constructed, and the subsequent search or selection mechanism. These systems are designed to improve final recovery, whereas our experiments isolate a narrower diagnostic: whether LLM-generated candidates produce paired gains beyond a fixed semantics-free library when data, fitting, and search budget are held constant.

Final recovery nevertheless aggregates several capability sources. Removing an agent, retrieval component, or search module in a conventional ablation can change the available candidates and the optimization budget at the same time, so the resulting accuracy difference cannot be attributed uniquely to scientific knowledge encoded in the language model. Scientific context may improve candidate recall without producing any structure absent from a fixed library, while retrieval may expand the pool using knowledge external to the model parameters. Both effects are useful, but they answer different evaluation questions. We therefore treat candidate origin, semantic access, and fixed-library coverage as separate experimental factors rather than as a single ``LLM versus non-LLM'' comparison.

\section{Experimental Methods}

The experiments use 129 LSR-Synth tasks from a local snapshot of LLM-SRBench, spanning four domains: population growth, physical oscillations, chemical reactions, and material relationships. Each task contains 4,000 training points, 500 in-distribution (ID) test points, and 500 out-of-distribution (OOD) test points, together with variable names, variable meanings, and a domain description \cite{shojaee2025llmsrbench,shojaee2025llmsrbenchdata}. In the main experiments, a fixed seed is used to sample up to 800 points from the training set for candidate selection and parameter fitting. The two test sets are reserved exclusively for final evaluation. Ground-truth expressions are not used in prompting, candidate generation, or fitting. They are used only for post hoc evaluation and, in the matched operator-knockout experiments in Section 3, to label the target operator families.
The main experiments evaluate three conditions—BANK, LLM, and UNION—separately on each task. All three use the same training subsample, fitting procedure, candidate-selection rules, and stopping criteria; they differ only in the candidate lists available to the search procedure. BANK uses fixed candidates constructed without access to task semantics, LLM uses candidates proposed by the language model, and UNION uses the union of both sets. We additionally consider two forms of language input, L1 and L2, and assess robustness using a second publicly sourced vocabulary, classical symbolic regression methods, and controlled library-ablation experiments. We first describe candidate construction and the paired search procedure, and then define the evaluation metrics separately.

\subsection{Candidate Construction and Paired Search}

Fixed candidates are constructed using two operator configurations drawn from public sources. The main experiments use BANK-LaSR, based on the PySR operator set reported in LaSR \cite{grayeli2024lasr}: binary addition, subtraction, multiplication, division, and exponentiation, together with the unary sine, cosine, exponential, logarithmic, and square-root operators. BANK-SRBench-PySR is derived from the official PySR wrapper in SRBench, with cosine and unrestricted exponentiation removed, and serves as a replication using a second full library \cite{lacava2021srbench,cavalab2021srbenchsoftware}. Both vocabularies are expanded over the task variables using the same shallow construction procedure. For each variable, the expansion includes first-, second-, and third-order powers, permitted unary transformations, and a limited set of cross-variable products. Logarithmic terms are written as \(\log(|v|+a)\), square-root terms as \(\sqrt{|v|}\), and unrestricted power terms as \(|v|^a\). Tangent, hyperbolic functions, inverse trigonometric functions, and arbitrary-depth nesting are excluded. After deduplication, BANK-LaSR contains 10, 30, and 60 candidates for tasks with one, two, and three variables, respectively; the corresponding counts for BANK-SRBench-PySR are 8, 24, and 48.

In each run, the LLM proposes 14 atomic terms without leading coefficients, after which the program generates up to 20 products involving distinct state variables. Under L1, the model receives the original variable names, domain description, training-data ranges, and structural features computed from the data. L2 retains the same numerical information and structural features but replaces the variables with anonymous symbols such as \(x_1\) and \(x_2\) and removes the domain description; the returned expressions are subsequently mapped back to the original variables. This procedure produces three candidate pools: BANK contains only BANK-LaSR candidates; LLM contains only the language-model-generated atomic terms and their products; and UNION is the deduplicated union of the two sets. The same search procedure is run from scratch on each candidate pool, rather than fitting BANK and LLM separately and then selecting whichever output performs better.

\begin{algorithm}[t]
\caption{Paired Search for BANK, LLM, and UNION}
\label{alg:algorithm1}
\begin{algorithmic}[1]
\REQUIRE Training set \(\mathcal{D}\), variables \(\mathcal{V}\), language condition \(c\), and fixed vocabulary \(\mathcal{B}\)
\ENSURE Predictive expressions for BANK, LLM, and UNION
\STATE Sample at most \(800\) points from \(\mathcal{D}\) using a fixed seed
\STATE \(\mathcal{P}_{\mathrm{BANK}}\leftarrow
\operatorname{Dedup}(\operatorname{ShallowExpand}(\mathcal{B},\mathcal{V}))\)
\STATE \(\mathcal{P}_{\mathrm{LLM}}\leftarrow
\operatorname{LLM}(c,\text{ranges},\text{structural features})\)
\STATE Add up to \(20\) cross-state products to \(\mathcal{P}_{\mathrm{LLM}}\), then deduplicate
\FOR{each \(\mathcal{P}\in
\{\mathcal{P}_{\mathrm{BANK}},\mathcal{P}_{\mathrm{LLM}},
\mathcal{P}_{\mathrm{BANK}}\cup\mathcal{P}_{\mathrm{LLM}}\}\)}
    \STATE \(\mathcal{S}\leftarrow\{\text{all first-order terms in }\mathcal{V}\}\)
    \FOR{\(r=1,\ldots,5\)}
        \FOR{each \(t\in\mathcal{P}\setminus\mathcal{S}\)}
            \STATE Tentatively add \(t\) and jointly re-estimate all parameters
        \ENDFOR
        \STATE \(t^\star\leftarrow
        \arg\min_{t\in\mathcal{P}\setminus\mathcal{S}}
        \operatorname{NMSE}_{\mathrm{train}}(\mathcal{S}\cup\{t\})\)
        \IF{\(t^\star\) reduces the current NMSE by less than \(0.1\%\)}
            \STATE \textbf{break}
        \ENDIF
        \STATE \(\mathcal{S}\leftarrow\mathcal{S}\cup\{t^\star\}\)
        \IF{\(\operatorname{NMSE}_{\mathrm{train}}(\mathcal{S})<10^{-10}\)}
            \STATE \textbf{break}
        \ENDIF
    \ENDFOR
    \STATE Save the jointly fitted expression defined by \(\mathcal{S}\)
\ENDFOR
\end{algorithmic}
\end{algorithm}

Each candidate term is assigned an independent leading coefficient. Internal parameters—including sinusoidal frequencies, exponential rates, logarithmic shifts, and unrestricted power exponents—are re-estimated jointly with all existing parameters whenever a new term is tentatively added. Nonlinear least squares uses 10 deterministic initializations, with a maximum of 5,000 function evaluations per run. For a given task, the three candidate pools share the same training subsample, five-round limit, initializations, and stopping rules. Because UNION contains more candidates, however, it performs more trial additions. BANK-SRBench-PySR is run separately using the same algorithm. Modifications for the polynomial-only library, matched operator knockouts, SINDy, and genetic programming are described individually in Section 3.

\subsection{Evaluation Metrics and Statistics}

\paragraph{Scope of evaluation.} We report in-distribution numerical accuracy, strict joint recovery on ID and OOD data, symbolic accuracy, and paired differences between BANK and UNION. For every reported proportion, success is first determined as a binary outcome for each task and then aggregated across the task set.
\paragraph{In-distribution numerical accuracy.} The primary comparison follows the public implementation of SR-Scientist \cite{xia2026srscientist,xia2026srscientistsoftware}. At a test point with a nonzero ground-truth value, the prediction is counted as correct if its relative error is strictly below a threshold \(\tau\). When the ground-truth value is numerically zero, an absolute-error tolerance of \(10^{-9}\) is used instead. Test points for which no finite prediction can be obtained are counted as incorrect:
\begin{equation}
c_{ij}(\tau)
=
\mathbf{1}\!\left[
\frac{\left|\hat{y}_{ij}-y_{ij}\right|}
     {\left|y_{ij}\right|}
< \tau
\right],
\quad y_{ij}\neq 0 .
\label{eq:nonzero-accuracy}
\end{equation}
\begin{equation}
c_{ij}(\tau)
=
\mathbf{1}\!\left[
\left|\hat{y}_{ij}\right|
\leq 10^{-9}
\right],
\quad y_{ij}=0 .
\label{eq:zero-accuracy}
\end{equation}

A task is assigned \(\operatorname{Acc}_i(\tau)=1\) only if the proportion of correctly predicted test points is strictly greater than \(95\%\); otherwise, \(\operatorname{Acc}_i(\tau)=0\). \(\operatorname{Acc}_{0.01}\) uses \(\tau=0.01\), whereas \(\operatorname{Acc}_{0.001}\) uses \(\tau=0.001\). Both metrics evaluate in-distribution function values and do not require the predicted and ground-truth expressions to share the same structure.
\paragraph{Strict joint recovery on ID and OOD data.} For any test set \(D\), the normalized mean squared error is defined as
\begin{equation}
\operatorname{NMSE}(D)
=
\frac{
\sum_{j\in D}(\hat{y}_j-y_j)^2
}{
\sum_{j\in D}\bigl(y_j-\bar{y}(D)\bigr)^2
}.
\end{equation}
A task is counted as strictly recovered only if the predicted expression produces finite values at every point in both test sets and satisfies \(\operatorname{NMSE}(\mathrm{ID}) < 10^{-8}\) and \(\operatorname{NMSE}(\mathrm{OOD}) < 10^{-8}\). Compared with the accuracy metrics, this criterion places greater emphasis on functional behavior outside the fitting interval, although it still does not constitute proof of symbolic identity.
\paragraph{Paired statistics.} For any task-level criterion, let \(I_i(M,c)\) indicate whether method \(M\) succeeds on task \(i\) under condition \(c\), and let \(N\) denote the number of tasks. We use the following statistics:
\begin{equation}
A(M,c)=\frac{1}{N}\sum_i I_i(M,c).
\end{equation}
\begin{equation}
\Delta_{\mathrm{prior}}(c)
=
\sum_i
\left[
I_i(\mathrm{UNION},c)-I_i(\mathrm{BANK},c)
\right].
\end{equation}
\begin{equation}
\Delta_{\mathrm{sem}}
=
A(\mathrm{LLM},\mathrm{L1})
-
A(\mathrm{LLM},\mathrm{L2}).
\end{equation}
Here, \(\Delta_{\mathrm{prior}}\) is the net number of additional successful tasks after task-level pairing, rather than the difference between two independently computed mean scores. Because the selection path of forward greedy search is not guaranteed to be monotonic as the candidate pool expands, UNION may occasionally select a different candidate at an early stage. Accordingly, \(\Delta_{\mathrm{prior}}\) is allowed to take a small negative value.
\paragraph{Symbolic accuracy (SA).} We use the ten-replicate judge prompt reported by SR-Scientist \cite{xia2026srscientist}, but adopt a stricter unanimity rule rather than a subsequent disagreement-review stage. All evaluations use one fixed prompt template; only the task-specific ground-truth expression \(A\) and predicted expression \(B\) are substituted. GPT-OSS-120B \cite{openai2025gptoss} independently evaluates each resulting expression pair 10 times. A task is assigned \(\operatorname{Sym}_i=1\) only if all 10 judgments return Yes; a unanimous No, any mixed vote, or an incomplete set of judgments is assigned \(\operatorname{Sym}_i=0\). The evaluation asks whether there exists a set of free constants \(\theta\), belonging only to the predicted expression, such that the predicted skeleton \(\hat{f}_i(\mathbf{x};\theta)\) is equivalent to the fixed ground-truth expression \(f_i^*(\mathbf{x})\). Physical parameters appearing in the ground-truth expression may not be reassigned, whereas leading coefficients on superfluous terms in the predicted expression may be set to zero. Symbolic accuracy is computed as:
\begin{equation}
\operatorname{SA}
=
\frac{1}{N}\sum_i \operatorname{Sym}_i.
\end{equation}
The main results are reviewed under a conservative identity criterion requiring equivalence over the full domain. This metric evaluates whether the predicted skeleton can represent the ground-truth structure; it does not imply that every parameter obtained in the current fit is individually correct. The deterministic rules used by LaSR and SRBench are included only as metric-sensitivity checks and do not contribute to the SA values reported in the main text \cite{grayeli2024lasr,lacava2021srbench,cavalab2021srbenchsoftware}.

\section{Experimental Results and Analysis}
This section first compares existing methods using the official LSR-Synth metrics and then investigates the factors underlying the main comparison through candidate-source, semantic-blinding, and library-ablation experiments. The main comparison reports \(\operatorname{Acc}_{0.01}\), \(\operatorname{Acc}_{0.001}\), and symbolic accuracy (SA).
The literature values are taken from the results reported by SR-Scientist on the same 129 tasks \cite{xia2026srscientist}. The original sources of the baseline methods are provided in \cite{landajuela2022udsr,cranmer2023pysr,grayeli2024lasr,shojaee2025llmsr,xia2026srscientist}, and only the best publicly reported value for each metric is retained for each method. All multi-seed main results use the same fixed seeds and all conditions are paired within each task and seed. Classical SINDy is run only once, and the confidence interval for the matched knockout experiment is clustered at the task level.

\subsection{Main Comparison with Existing Methods}
\label{subsec:main_comparison}

\begin{table}[t]
\centering
\caption{Comparison on LSR-Synth. Our values are mean \(\pm\) population standard deviation over four paired seeds; literature values are the best publicly reported values.}
\label{tab:main_results}
{\small
\setlength{\tabcolsep}{3pt}
\begin{tabular}{lccc}
\toprule
Method &
\shortstack{Acc\textsubscript{0.01}\\(\%)} &
\shortstack{Acc\textsubscript{0.001}\\(\%)} &
\shortstack{SA\\(\%)} \\
\midrule
uDSR
& 29.46
& 12.40
& 0.77 \\

PySR
& 29.46
& 14.47
& 4.65 \\

LaSR
& 16.02
& 10.08
& --- \\

LLM-SR
& 41.08
& 18.09
& 5.43 \\

SR-Scientist
& 63.57
& 49.35
& 7.75 \\

\textbf{BANK-LaSR (ours)}
& \textbf{77.71}\(\boldsymbol{\pm}\)\textbf{1.49}
& \textbf{62.60}\(\boldsymbol{\pm}\)\textbf{0.84}
& \textbf{19.77}\(\boldsymbol{\pm}\)\textbf{1.16} \\
\bottomrule
\end{tabular}
}

\end{table}

For each metric, Table~\ref{tab:main_results} retains only the best publicly reported result in the literature; SA was not reported for LaSR. BANK-LaSR outperforms these results on both numerical-accuracy metrics and SA, indicating that a fixed candidate library with no access to domain semantics can already achieve a high score on LSR-Synth. This comparison does not imply that BANK has superior general-purpose symbolic regression capabilities, because the search representations and computational budgets are not identical. Rather, it shows that further analysis is required to determine whether the current high scores genuinely arise from language-model priors. Our results are reported as the mean and population standard deviation over four fixed, paired seeds. Our SA values use the conservative 10-of-10 unanimity rule defined above.

\subsection{Candidate Sources and Operator-Library Coverage}
\begin{table*}[!t]
\centering
\caption{Paired candidate-source and library-coverage results (\%). Values are mean \(\pm\) population standard deviation over four paired seeds.}
\label{tab:candidate_results}
{\small
\setlength{\tabcolsep}{5pt}
\begin{tabular}{lcccc}
\toprule
Condition &
Acc\textsubscript{0.01} &
Acc\textsubscript{0.001} &
Strict ID+OOD &
SA \\
\midrule
BANK-LaSR
& 77.71\(\pm\)1.49
& 62.60\(\pm\)0.84
& 39.53\(\pm\)2.26
& 19.77\(\pm\)1.16 \\

LLM-L1
& 49.81\(\pm\)1.85
& 30.81\(\pm\)2.22
& 13.76\(\pm\)1.49
& 8.14\(\pm\)0.39 \\

LLM-L2
& 41.09\(\pm\)1.98
& 22.09\(\pm\)1.78
& 6.59\(\pm\)1.69
& 2.71\(\pm\)0.87 \\

UNION-LaSR-L1
& 77.33\(\pm\)0.34
& 62.79\(\pm\)1.45
& 39.34\(\pm\)1.15
& 18.80\(\pm\)1.49 \\

UNION-LaSR-L2
& 77.71\(\pm\)1.77
& 63.76\(\pm\)1.49
& 40.89\(\pm\)1.93
& 20.16\(\pm\)1.10 \\

BANK-Poly
& 27.52\(\pm\)0.39
& 8.91\(\pm\)0.67
& 4.65\(\pm\)0.00
& 1.55\(\pm\)0.00 \\

UNION-Poly-L1
& 58.14\(\pm\)1.64
& 36.24\(\pm\)1.85
& 16.67\(\pm\)1.16
& 10.08\(\pm\)0.00 \\

UNION-Poly-L2
& 49.81\(\pm\)2.15
& 29.46\(\pm\)1.90
& 11.82\(\pm\)0.64
& 5.04\(\pm\)0.87 \\
\bottomrule
\end{tabular}
}

\end{table*}

\begin{figure*}[!t]
\centering
\includegraphics[width=\textwidth]{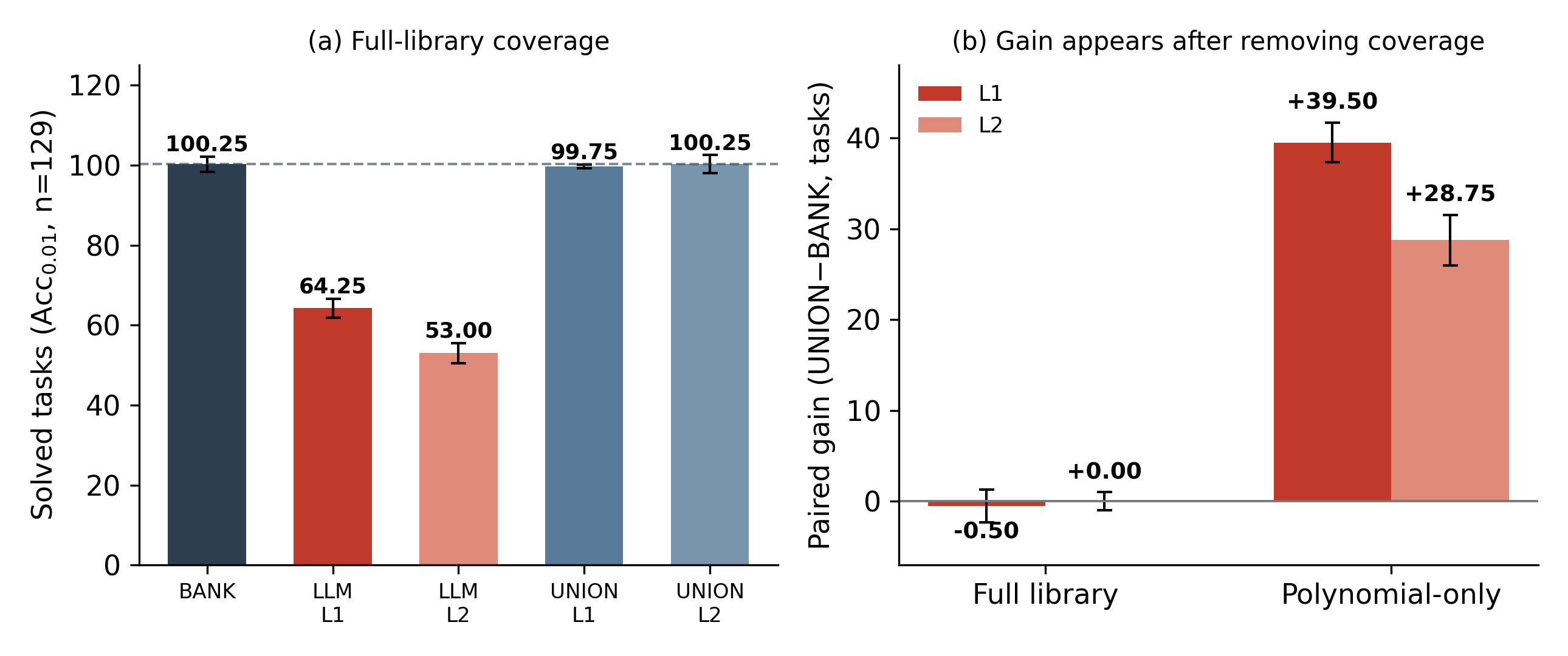}
\caption{Candidate-source comparison under full-library and polynomial-only coverage using \(\operatorname{Acc}_{0.01}\). Bars show either the mean number of solved tasks or the paired UNION--BANK gain across four fixed seeds; error bars show the population standard deviation across seeds.}
\label{fig:candidate_sources}
\end{figure*}

\begin{table*}[t]
\centering
\caption{Paired four-cell counts for the BANK-LaSR and LLM standalone arms.}
\label{tab:paired_counts}
{\small
\setlength{\tabcolsep}{3.5pt}
\begin{tabular}{llrrrrrrrr}
\toprule
& &
\multicolumn{2}{c}{Acc\textsubscript{0.01}} &
\multicolumn{2}{c}{Acc\textsubscript{0.001}} &
\multicolumn{2}{c}{Strict ID+OOD} &
\multicolumn{2}{c}{SA} \\
\cmidrule(lr){3-4}
\cmidrule(lr){5-6}
\cmidrule(lr){7-8}
\cmidrule(lr){9-10}
BANK & LLM &
L1 & L2 &
L1 & L2 &
L1 & L2 &
L1 & L2 \\
\midrule
Correct
& Correct
& 254 & 210
& 139 & 99
& 65 & 27
& 16 & 2 \\

Correct
& Incorrect
& 147 & 191
& 184 & 224
& 139 & 177
& 86 & 100 \\

Incorrect
& Correct
& 3 & 2
& 20 & 15
& 6 & 7
& 26 & 12 \\

Incorrect
& Incorrect
& 112 & 113
& 173 & 178
& 306 & 305
& 388 & 402 \\
\bottomrule
\end{tabular}
}

\end{table*}

The main comparison shows that a simple fixed library can achieve high scores, but it does not by itself establish whether the language model provides additional benefits. We therefore fix the seeds to 0, 1, 2, and 3 instead of randomly selecting different seeds for different methods. For a given task and seed, all conditions share the same training subsample, fitter, and search budget. The L1 and L2 candidates generated under that seed are also reused in the full-library and polynomial-library settings, respectively. Consequently, differences in the results arise from the candidate source rather than from one condition fortuitously receiving more favorable data. Tables~\ref{tab:main_results} and~\ref{tab:candidate_results} use exactly the same four seeds and BANK expressions. Each row of Table~\ref{tab:candidate_results} corresponds to a distinct candidate pool. BANK-LaSR does not use a language model and searches only the fixed candidates expanded from the LaSR operator vocabulary. LLM-L1 searches only candidates proposed by the language model when given the true variable names and domain description. LLM-L2 also uses only language-model candidates, but the variables are anonymized and the domain labels are removed. UNION-LaSR-L1 augments BANK-LaSR with the L1 candidates, whereas UNION-LaSR-L2 augments the same BANK-LaSR pool with L2 candidates that contain no scientific semantics. Both UNION conditions then run the same greedy search from scratch, rather than selecting the better of two previously obtained solutions. The paired comparison between BANK-LaSR and UNION-LaSR-L2 therefore addresses a specific question: without scientific context, can generic mathematical candidates proposed by the language model enable the full fixed library to solve additional tasks?

The final three rows deliberately introduce gaps in library coverage. BANK-Poly weakens the fixed library to first- through fourth-order powers of each variable and a limited set of variable products. UNION-Poly-L1 and UNION-Poly-L2 then add the LLM candidates generated under the two respective language conditions. We do not include a separate LLM-Poly condition, because the LLM-only arm does not use the fixed library and is therefore identical to LLM-L1 or LLM-L2 above. The two \(\operatorname{Acc}\) metrics are evaluated only on in-distribution test points. By contrast, strict joint recovery requires the same saved expression to produce finite values on all ID and OOD points and to satisfy \(\operatorname{NMSE}(\mathrm{ID}) < 10^{-8}\) and \(\operatorname{NMSE}(\mathrm{OOD}) < 10^{-8}\). We do not evaluate OOD performance in isolation, because an expression that already fails within the fitting range should not be regarded as extrapolating successfully. For SA, GPT-OSS-120B independently evaluates each parameterized expression ten times, and only a 10-of-10 Yes result is counted as correct. Mixed votes and unanimous No results are both scored as failures without manual relabeling. All values are the mean and population standard deviation over seeds 0, 1, 2, and 3.

\paragraph{Overall Results.} With the full LaSR library, both UNION conditions perform similarly to BANK, whereas the LLM-only arms perform substantially worse. When the fixed library is weakened to a polynomial library, adding LLM candidates yields clear improvements across all four metrics. LLM-L1 also consistently outperforms LLM-L2, indicating that scientific context helps the model recall useful candidates. However, this benefit translates into gains over BANK only when the candidates fill gaps in the fixed library and are actually selected by the search procedure.

\paragraph{Out-of-Distribution and Symbolic Results.} Adding the OOD requirement lowers the absolute success rate of every condition, but it does not reverse the overall relationship: performance changes little for the full library, whereas the weakened library benefits substantially from LLM candidates. SA provides the same contrast. With the full library, UNION-L1 performs slightly worse than BANK, while UNION-L2 exhibits only a small fluctuation. With the polynomial library, adding LLM candidates increases SA from 1.55\% to 10.08\% under L1 or 5.04\% under L2. The conclusion is therefore not an artifact of any single numerical threshold. SA remains judge-dependent, and the unanimity rule deliberately treats all disputed cases as failures rather than resolving them in favor of either expression.

\paragraph{Matched knockout.}
After operators required by the ground-truth equations are removed, \(\operatorname{Acc}_{0.01}\) is lower than in the control condition, in which an equal number of irrelevant operators are removed. However, the \(95\%\) confidence interval in the current 15-task pilot experiment still includes zero. This result can therefore be treated only as directional evidence and cannot be presented as confirmation of the coverage mechanism. A formal test with an expanded sample remains to be completed.

\subsection{Paired Decomposition of BANK and LLM Success Sets}
\label{subsec:paired_decomposition}

Table~\ref{tab:candidate_results} compares final average scores but does not reveal whether the two standalone arms solve the same or different tasks. Table~\ref{tab:paired_counts} therefore pairs BANK-LaSR with the LLM condition on the same task and seed and partitions the 516 task--seed instances into four categories: both correct, BANK only, LLM only, and both incorrect. The two \(\operatorname{Acc}\) metrics and strict \(\mathrm{ID{+}OOD}\) recovery use the corresponding task-level numerical labels, whereas SA uses the final expression-level symbolic labels. This decomposition first determines whether the LLM finds solutions that BANK does not and then examines whether those solutions are actually retained by UNION.

Table~\ref{tab:paired_counts} reveals three general patterns. First, under the more permissive \(\operatorname{Acc}_{0.01}\) threshold, the two candidate types overlap extensively. Of the 257 L1 successes, 254 are also solved by BANK; of the 212 L2 successes, 210 overlap with BANK. Thus, the LLM contributes only three and two unique successes, respectively. Second, when the threshold is tightened, OOD performance is required, or SA is considered, the number of LLM-only successes increases but remains far smaller than the number of BANK-only successes. For SA, for example, L1 and L2 yield 26 and 12 LLM-only structural forms, respectively, compared with 86 and 100 BANK-only structural forms. Third, the LLM generally succeeds more often in the L1 columns than in the L2 columns. This finding indicates that domain context improves candidate recall, but it does not change the fact that the full fixed library covers more tasks.

Considering the full and weakened libraries together makes the pattern even clearer. With the full LaSR library, the LLM produces only a limited number of distinct successful answers, and the final UNION score generally does not exceed that of BANK. Once exponential, logarithmic, trigonometric, and free-power terms are removed from the polynomial library, the number of LLM-only successes increases substantially, and most of them translate into UNION successes. Specifically, under \(\operatorname{Acc}_{0.01}\), UNION-Poly-L1 retains 141 of the 144 LLM-only successes; under strict \(\mathrm{ID{+}OOD}\) recovery, it retains 60 of 61. For SA, all 42 LLM-only structural forms under L1 and all 13 under L2 are retained. The paired statistics therefore primarily support the following general pattern: LLM candidates yield stable complementary gains only when they fill genuine gaps in the fixed library.

\begin{table*}[t]
\centering
\caption{Representative LLM-assisted symbolic regression methods organized by intervention stage. The comparison describes mechanisms rather than ranking performance.}
\label{tab:method_map}
{\small
\setlength{\tabcolsep}{4pt}
\begin{tabular}{
>{\raggedright\arraybackslash}p{1.05in}
>{\raggedright\arraybackslash}p{1.45in}
>{\raggedright\arraybackslash}p{1.85in}
>{\raggedright\arraybackslash}p{1.80in}}
\toprule
Method & Information source & Candidate construction & Search or selection \\
\midrule
DrSR \cite{wang2025drsr}
& Data-derived structural features
& LLM-generated equation proposals
& Iterative feedback and refinement \\
PG-SR \cite{xiao2026pgsr}
& Executable scientific constraints
& Constraint-checked equations
& Prior-guided evolution and refinement \\
SR-LLM \cite{guo2025srllm}
& Retrieved knowledge and prior runs
& Retrieved symbolic primitives
& Reinforcement-learning-based assembly \\
IGSR \cite{saveliev2026igsr}
& Task data and influence feedback
& LLM-generated basis terms
& Tree search and influence-guided pruning \\
\bottomrule
\end{tabular}
}

\end{table*}

The few full-library tasks solved only by the LLM do not consistently remain solved by UNION because UNION reruns the search over the combined pool rather than selecting the better standalone answer. Candidate competition can therefore both discard standalone successes and create new successes through interactions among candidates. Consequently, UNION--BANK should be interpreted as an end-to-end gain rather than a pure measure of candidate novelty.

\section{Discussion and Implications}
\label{sec:discussion}

\paragraph{What LSR-Synth measures.}
LSR-Synth reduces direct formula memorization by combining and modifying scientific mechanisms, so its targets cannot simply be copied as complete equations from the public literature. This is an important improvement over collections dominated by canonical formulas. However, novelty of the complete equation and novelty of the search space are different properties. To keep the generated tasks scientifically plausible, numerically stable, and solvable, their components still rely mainly on familiar powers, rational functions, exponentials, trigonometric terms, and low-order interactions. A target can therefore be new as a whole while remaining reachable from a conventional vocabulary that never reads the variable names or domain description.

The contrast between the full and weakened libraries provides the central evidence for this distinction. BANK-LaSR already covers many required component structures, and its successful set contains almost all LLM successes at the 1\% threshold. In this setting, adding LLM candidates changes the final score only slightly and non-monotonically. When the fixed library is restricted to polynomials, by contrast, LLM candidates restore missing non-polynomial structures and produce large, consistent gains under both numerical and symbolic criteria. Semantic context improves the LLM-only arm in both settings, but that improvement becomes an end-to-end advantage over BANK only when it fills an actual coverage gap. Aggregate LSR-Synth accuracy can thus demonstrate recovery of unseen complete equations; without a semantics-free control, it cannot by itself identify how much of that recovery requires scientific priors beyond a conventional search space.

\paragraph{Implications for benchmark design.}
Future benchmarks can preserve complete-formula novelty while separately controlling semantics-free reachability. Before release, generated tasks can be stress-tested with documented fixed libraries and prescribed budgets. Tasks readily solved by these controls can be labeled as closed-vocabulary recovery, whereas an open-vocabulary split can reserve operator families, named mechanisms, or higher-order interactions not disclosed to the evaluated method. Reporting BANK, LLM, and UNION on the same task--seed pairs would then distinguish candidate recall from effective boundary expansion. Matched target-operator and sham knockouts provide an additional diagnostic: the former removes structures used by the target, whereas the latter changes library size without removing required structures. These controls make the intended claim auditable without requiring every task to be unsolvable by classical search.

\paragraph{Scope and limitations.}
The evidence does not support a claim that language models are generally unnecessary for symbolic regression. It is limited to the current LSR-Synth snapshot, the tested vocabularies, and the prescribed search budgets. The matched knockout covers only 15 tasks and its \(\operatorname{Acc}_{0.01}\) confidence interval includes zero; the cross-backbone, GP, and selection diagnostics use historical single runs or task subsets; strict ID+OOD recovery depends on predefined extrapolation intervals; and SA depends on expression parameterization and judge rules. The unanimity rule further biases SA downward by assigning every mixed judgment to failure; it avoids optimistic manual relabeling but is not a model-independent definition of symbolic equivalence. The narrower conclusion is that commonly used operators already cover many targets in this benchmark, so aggregate accuracy alone is insufficient evidence that an LLM supplied indispensable scientific knowledge.

\section{Conclusion}
\label{sec:conclusion}

Using fixed libraries, semantic blinding, candidate unions, and library interventions, we distinguish novelty of a complete formula from novelty of its search space. A semantics-free fixed library reaches a large fraction of LSR-Synth, while LLM candidates yield stable complementary gains mainly after a genuine structural gap is introduced. The result does not diminish the value of anti-memorization benchmarks; it narrows the claim supported by their aggregate scores.

Evaluations intended to demonstrate scientific priors should pair complete-expression novelty with documented reachability controls. Paired BANK, LLM, and UNION results then distinguish candidate recall from effective search-boundary expansion.

\bibliography{references}
\bibliographystyle{icml2026}

\appendix

\section{Experimental Implementation Details}
\label{app:implementation}

\subsection{Data Snapshot and Randomness}
\label{app:data_snapshot}

The experiments use the LSR-Synth split from a local LLM-SRBench snapshot, containing 129 tasks: 24 biological, 44 physical, 36 chemical, and 25 materials-science tasks. An earlier version of the LLM-SRBench paper reported 128 tasks, including 43 physical tasks; all tables and task-level pairings in this study are based on the 129-task snapshot actually used in our experiments. Each task contains 4,000 training points, 500 in-distribution (ID) test points, and 500 out-of-distribution (OOD) test points. The training set is used only for candidate selection and parameter estimation, and neither test set participates in the search.

The main experiments use fixed seeds 0, 1, 2, and 3. For each task--seed combination, the same subsample of at most 800 training points is used by BANK, LLM, and UNION, as well as by the full and weakened libraries, yielding 516 paired units. The seed does not change the ground-truth equation or either test set. Candidate generation through the server API does not provide a verifiably deterministic seed; we therefore save the atomic terms produced by every generation call and do not invoke the model again when recomputing the evaluation metrics.

\subsection{Fixed Candidate Libraries}
\label{app:fixed_libraries}

The fixed libraries generate candidates mechanically from the input variables without accessing variable meanings, domain descriptions, or ground-truth equations. The operator set for BANK-LaSR is taken from Figure~7 of LaSR, whereas BANK-SRBench-PySR is based on commit \texttt{5b456d5} of the official PySR wrapper in SRBench. We reuse only the publicly documented operator sets, not the genetic-search procedures of the original methods.

\begin{table*}[t]
\centering
\caption{Construction of the fixed candidate libraries.}
\label{tab:fixed_libraries}
{\small
\setlength{\tabcolsep}{1.4pt}
\begin{tabular}{
>{\raggedright\arraybackslash}p{1.15in}
>{\raggedright\arraybackslash}p{2.05in}
>{\raggedright\arraybackslash}p{2.55in}
>{\centering\arraybackslash}p{0.85in}
}
\toprule
\multicolumn{1}{l}{Candidate library} &
\multicolumn{1}{c}{Single-variable candidates} &
\multicolumn{1}{c}{Cross-variable candidates} &
\multicolumn{1}{c}{Size (1/2/3)} \\
\midrule

BANK-LaSR &
\(v\), \(v^2\), \(v^3\), \(\sin(av)\), \(\cos(av)\),
\(\exp(\pm av)\), \(\log(|v|+a)\), \(\sqrt{|v|}\), and \(|v|^a\) &
\(g(v_i)v_j\), where
\(g\in\{v,v^2,\sin(av),\exp(-av),|v|^a\}\) &
10, 30, 60 \\

\addlinespace

BANK-SRBench-PySR &
BANK-LaSR without \(\cos(av)\) and \(|v|^a\) &
The same expansion rule, excluding interaction terms containing free
powers &
8, 24, 48 \\

\addlinespace

BANK-Poly &
First- through fourth-order powers of each variable &
\(v_i v_j\) and \(v_i^2v_j\) &
Variable-dependent \\

\bottomrule
\end{tabular}
}

\end{table*}

The absolute-value operations in the logarithmic, square-root, and free-power terms are used only to protect the function domain and are not treated as independent operators. None of the three libraries includes fragments taken from the ground-truth equations. BANK-Poly is a positive control designed to introduce explicit coverage gaps; it uses the same training subsamples, selector, and fitter as the full libraries.

\subsection{Search, Fitting, and Candidate Sources}
\label{app:search_fitting}

\begin{table}[t]
\centering
\caption{Search and fitting configurations.}
\label{tab:search_configuration}
{\small
\setlength{\tabcolsep}{2pt}
\begin{tabular}{
>{\raggedright\arraybackslash}p{0.86in}
>{\centering\arraybackslash}p{0.94in}
>{\centering\arraybackslash}p{1.08in}}
\toprule
Setting & Main experiment & Matched knockout \\
\midrule
Training points & At most 800 & 300 \\
Greedy rounds & At most 5 & At most 5 \\
Acceptance threshold & At least 0.1\% NMSE reduction & Same \\
Nonlinear initializations & 10 & 4 \\
Function-call limit & 5,000 & 1,200 \\
Early-stopping error & \(10^{-10}\) & \(10^{-10}\) \\
\bottomrule
\end{tabular}
}

\end{table}

The search begins with all first-order input-variable terms. In each round, every unselected candidate is tentatively added in turn, and all leading coefficients and internal parameters of the existing and newly added terms are jointly re-estimated. The candidate producing the lowest training NMSE is then retained. Internal parameters in different atomic terms are indexed separately and do not share a common \(a\). The LLM conditions first generate 14 atomic terms and then add at most 20 products involving distinct state variables. BANK contains only fixed candidates, LLM contains only model-generated candidates, and UNION deduplicates the two candidate sets and reruns the search from scratch rather than selecting between the final expressions produced by the two standalone arms.

\section{Evaluation Protocol}
\label{app:evaluation}

\subsection{Numerical Metrics and Paired Statistics}
\label{app:numerical_metrics}

The definitions of \(\operatorname{Acc}_{0.01}\), \(\operatorname{Acc}_{0.001}\), and NMSE are provided in Section~2.2 of the main paper. The implementation applies strict relative-error thresholds to nonzero targets and an absolute tolerance of \(10^{-9}\) to numerical zeros; nonfinite predictions are incorrect, and strictly more than 95\% of points must pass. Strict ID+OOD recovery uses the same saved expression without refitting and requires finite predictions with NMSE below \(10^{-8}\) on both test sets.

The \(\pm\) values in Tables~1 and 2 of the main paper are population standard deviations of the success rates across the four fixed seeds, not confidence intervals. All BANK--LLM--UNION comparisons are paired using the same task, seed, and training subsample. Confidence intervals for the matched operator-knockout experiment are computed using task-clustered bootstrap sampling, such that the four seeds associated with the same task are always sampled as a single group.

\subsection{Symbolic Accuracy and the Unanimity Rule}
\label{app:symbolic_accuracy}

The ground-truth expression remains fixed during symbolic evaluation. Fitted floating-point coefficients, frequencies, scales, and floating-point exponents in the predicted expression are parameterized, whereas structural integer powers are retained. Only the task-defined state notations \texttt{P(t)}, \texttt{A(t)}, \texttt{v(t)}, and \texttt{x(t)} are converted back into variables; mathematical functions such as \texttt{sin(t)} and \texttt{Abs(t)} are not rewritten. GPT-OSS-120B evaluates the expression pair in each scoring row ten times at a temperature of 0.7. The temperature is a replication assumption because SR-Scientist does not separately disclose the judge temperature.

We count an expression as symbolically correct only when all ten votes are Yes. Mixed votes and unanimous-No cases are assigned zero without manual relabeling. The identity criterion permits free parameters only in the predicted expression and does not permit reassignment of physical parameters in the ground-truth expression. Coefficients on additional predicted terms may be set to zero, while domain-protection forms with different domains are not automatically treated as equivalent. The archive contains \(129\) tasks \(\times\) four seeds \(\times\) eight candidate-source conditions, giving 4,128 task--seed--condition rows. A row is one scoring unit, not a separately designed prompt: the same template is filled with that row's ground-truth and predicted expressions. Every row receives its own independently sampled batch of ten judgments, giving 41,280 judge-model calls in total. Of the 4,128 rows, 445 are unanimous Yes, 2,870 are unanimous No, and 813 have mixed votes. The reported SA values use only the 445 unanimous-Yes rows as positives.

\section{Prompt Templates and Blinding Protocol}
\label{app:prompts}

\subsection{Candidate-Generation Prompt}
\label{app:candidate_prompt}

The system message is: \texttt{You propose atomic terms for symbolic regression. JSON only.} The user prompt template is shown below, where the fields enclosed in braces are populated at runtime using the training data and task metadata.

\begin{listing}[t]
\caption{Candidate-generation prompt template.}
\label{lst:candidate_prompt}
\begin{lstlisting}[numbers=none]
You are decomposing an ODE's right-hand side into ATOMIC building-block terms.
Variables: {inputs}
Target: y (= time-derivative of a state variable).
Domain: {hint}

Data profile:
{profile}

Structural facts from the data:
{invariants}

Do NOT write the whole formula. Instead propose a MENU of 14 candidate ATOMIC terms
(building blocks) that MIGHT appear as additive pieces of y. Cover a DIVERSE set of the
standard operator families used in symbolic regression: polynomial, rational, exponential,
logarithmic, trigonometric, power (including non-integer), and products of variables.

Rules for each atom:
- Write it WITHOUT a leading coefficient (the coefficient is added automatically).
- For an INNER constant (rate/frequency/exponent), use the symbol 'a' (one per atom).
- Must be SymPy-parseable, using only variables {inputs} and 'a'.

Reply JSON ONLY: {"atoms": ["...", "...", ...]}
\end{lstlisting}
\end{listing}

The prompt does not contain the ground-truth equation, ground-truth operator labels, or raw sample rows. The \texttt{profile} field reports only the ranges of the inputs and target. The \texttt{invariants} field contains features computed from the training data, potentially including robust equilibria, monotonic or saturating trends, asymptotic powers, parity, and prominent periodicity. Structural features that cannot be estimated reliably are omitted. The model temperature is 0.7, and the maximum output length is 900 tokens.

L1 and L2 differ only in the language context supplied to the model.

\begin{table}[t]
\centering
\caption{Differences between the L1 and L2 candidate-generation conditions.}
\label{tab:l1_l2_prompts}
{\small
\setlength{\tabcolsep}{2pt}
\begin{tabular}{
>{\raggedright\arraybackslash}p{0.74in}
>{\raggedright\arraybackslash}p{0.93in}
>{\raggedright\arraybackslash}p{1.23in}}
\toprule
Input field & L1 & L2 \\
\midrule
Variables &
Original identifiers, e.g., \texttt{t,P} &
Anonymous identifiers, e.g., \texttt{x1,x2} \\

Domain &
Original task description &
No domain information; predict derivative \(y\) from the listed inputs \\

Numerical ranges &
Retained &
Identical to L1 \\

Data-derived structural features &
Retained &
Rephrased using anonymous variables \\

Candidate count and operator instructions &
Identical across conditions &
Identical across conditions \\
\bottomrule
\end{tabular}
}

\end{table}

The L2 outputs are mapped back to the original variables using token-boundary matching. If JSON parsing fails, an expression cannot be parsed, or an undeclared variable is used, the corresponding atom is excluded from fitting. The full and polynomial libraries reuse the same L1 and L2 outputs, ensuring that the library-ablation experiment changes only the coverage of the fixed candidates.

\subsection{Symbolic-Judge Prompt}
\label{app:judge_prompt}

Symbolic evaluation uses the prompt from Figure~17 of SR-Scientist:

\begin{listing}[t]
\caption{Symbolic-judge prompt template.}
\label{lst:symbolic_judge_prompt}
\begin{lstlisting}[numbers=none]
Given the ground truth mathematical expression A and the hypothesis B,
determine if there exist any constant parameter values that would make the hypothesis
equivalent to the given ground truth expression.
Let's think step by step. Explain your reasoning and then provide the final answer as:
{ "reasoning": "Step-by-step analysis", "answer": "Yes/No" }
Ground Truth A: {ground_truth}
Hypothesis B: {parameterized_prediction}
\end{lstlisting}
\end{listing}

This prompt determines only whether the parameterized predicted skeleton can represent the fixed ground-truth expression; it does not determine whether every parameter estimated in the current fit is individually correct. The ten raw outputs and their aggregate vote labels should be released with the supplementary material but are not reproduced individually in the paper.

\section{Supplementary Results and Analysis}
\label{app:supplementary_results}

\subsection{Additional Results by Domain and Backbone}
\label{app:domain_backbone_results}

\paragraph{Results by Domain.} Table~\ref{tab:domain_results} aggregates the four paired runs by task domain.

\begin{table}[t]
\centering
\caption{Mean numbers of tasks solved under \(\operatorname{Acc}_{0.01}\) across four paired seeds. \(\Delta\) denotes UNION--BANK.}
\label{tab:domain_results}
{\small
\setlength{\tabcolsep}{1.5pt}
\begin{tabular}{lrrrrrr}
\toprule
Domain & \(N\) & BANK & L1 & L2 & \(\Delta\)L1 & \(\Delta\)L2 \\
\midrule
Biology   & 24  & 16.50  & 10.25 & 6.75  & \(-\)0.75 & \(-\)0.25 \\
Physics   & 44  & 28.25  & 7.25  & 4.75  & +0.25     & 0.00 \\
Chemistry & 36  & 32.75  & 28.00 & 24.25 & 0.00      & \(-\)0.25 \\
Materials & 25  & 22.75  & 18.75 & 17.25 & 0.00      & +0.50 \\
Total     & 129 & 100.25 & 64.25 & 53.00 & \(-\)0.50 & 0.00 \\
\bottomrule
\end{tabular}
}

\end{table}

The entries are the mean numbers of tasks solved under \(\operatorname{Acc}_{0.01}\) across the four seeds and may therefore be fractional. The improvement from L2 to L1 is distributed across multiple domains, whereas the net changes produced by UNION with the full library are both small and dispersed.

\paragraph{Backbone Check.} Table~\ref{tab:backbone_results} is based on previously saved single-run records for different candidate-generation backbones. These records are not part of the unified four-seed main experiment and do not contribute to the main-effect estimates. They are used only to examine whether the high coverage of the full library depends strongly on a particular language model.

\begin{table*}[t]
\centering
\caption{Single-run paired results across candidate-generation backbones. Gain/loss columns report task-level successes gained and lost by UNION relative to BANK.}
\label{tab:backbone_results}
{\small
\setlength{\tabcolsep}{2.5pt}
\begin{tabular}{lrrrrrrrr}
\toprule
Backbone & Tasks & BANK & LLM-L1 & LLM-L2 & UNION-L1 & UNION-L2 &
\shortstack{L1\\gain/loss} & \shortstack{L2\\gain/loss} \\
\midrule
Qwen3-Coder-480B & 129 & 99  & 69 & 57 & 100 & 98  & 2/1 & 0/1 \\
GPT-4.1          & 129 & 102 & 63 & 56 & 104 & 100 & 2/0 & 0/2 \\
GLM-4.6          & 128 & 101 & 68 & 62 & 98  & 101 & 1/4 & 1/1 \\
DeepSeek-V3.1    & 129 & 102 & 67 & 60 & 102 & 98  & 1/1 & 1/5 \\
\bottomrule
\end{tabular}
}

\end{table*}

The cross-backbone results are historical single runs and are used only as a directional check. The GLM record is missing one task, while the DeepSeek record retains the first complete entry for each task name. These runs use the historical 1\% test-relative-error criterion and cannot be combined with the four-seed main experiment to estimate variance.

\subsection{Candidate-Space Interventions and Additional Fixed Libraries}
\label{app:candidate_interventions}

The diagnostic set contains tasks that depend on exactly one target family among exponential, trigonometric, logarithmic, and fractional-power structures. We deterministically select the first task by name within each nonempty domain-by-operator-family stratum, yielding 15 tasks: BPG1, BPG12, BPG0; PO19, PO10, PO6, PO28; CRK22, CRK13, CRK35, CRK16; and MatSci13, MatSci20, MatSci19, MatSci0. Ground-truth expressions are used only to determine the intervention category and for post hoc scoring.

TARGET-KO removes the target family required by the ground-truth structure, whereas SHAM-KO removes an equal number of candidates that do not belong to the ground-truth structure. The three conditions share the same training points, fitting initializations, and search budget.

\begin{table*}[t]
\centering
\caption{Matched operator-family knockout results. The first three result columns report mean \(\pm\) population standard deviation over seeds 0--3. Confidence intervals use task-clustered bootstrap resampling.}
\label{tab:operator_knockout_seeds}
{\small
\setlength{\tabcolsep}{5pt}
\begin{tabular}{lrrrr}
\toprule
Metric & FULL (\%) & TARGET-KO (\%) & SHAM-KO (\%) &
\shortstack{SHAM--TARGET\\(percentage points)} \\
\midrule
Acc\textsubscript{0.01}
& \(63.33\pm7.45\) & \(53.33\pm0.00\) & \(63.33\pm7.45\)
& \(+10.0\), 95\% CI \([0.0,21.7]\) \\
Strict ID+OOD
& \(20.00\pm8.16\) & \(1.67\pm2.89\) & \(21.67\pm2.89\)
& \(+20.0\), 95\% CI \([6.7,36.7]\) \\
\bottomrule
\end{tabular}
}

\end{table*}

\begin{table}[t]
\centering
\caption{Matched knockout effects (SHAM--TARGET) by target operator family, in percentage points.}
\label{tab:operator_knockout_families}
{\small
\setlength{\tabcolsep}{2pt}
\begin{tabular}{lccc}
\toprule
Family & \(N\) & \(\Delta\)Acc & \(\Delta\)ID+OOD \\
\midrule
Exponential      & 4  & 0.0 pp   & +25.0 pp \\
Trigonometric    & 4  & +6.2 pp  & +12.5 pp \\
Logarithmic      & 3  & +16.7 pp & +16.7 pp \\
Fractional power & 4  & +18.8 pp & +25.0 pp \\
\bottomrule
\end{tabular}
}

\end{table}

This experiment tests only whether removing structures required by the ground truth is more harmful than removing an equal number of irrelevant candidates. The stratified samples are too small to support comparisons of the general difficulty of different operator families.

\begin{table}[t]
\centering
\caption{Additional fixed-library controls. BANK-SRBench-PySR reports mean \(\pm\) population standard deviation over four fixed seeds; SINDy/STLSQ has one run and therefore no error term.}
\label{tab:additional_fixed_libraries}
{\small
\setlength{\tabcolsep}{1.8pt}
\begin{tabular}{
>{\raggedright\arraybackslash}p{1.22in}
>{\centering\arraybackslash}p{0.58in}
>{\centering\arraybackslash}p{0.62in}
>{\centering\arraybackslash}p{0.62in}}
\toprule
Control & Runs & Acc\textsubscript{0.01} & ID+OOD \\
\midrule
BANK-SRBench-PySR &
\(129\times4\) &
\(69.96\pm2.22\%\) &
\(23.64\pm0.87\%\) \\

SINDy/STLSQ fixed-frequency library &
\(129\times1\) &
\(55.81\%\) (single run) &
\(21.71\%\) (single run) \\
\bottomrule
\end{tabular}
}

\end{table}

The narrower BANK-SRBench-PySR vocabulary still provides high coverage.
The SINDy condition does not fit internal function scales and is included only
as a classical fixed-library reference.

The stronger-GP, fixed-pool selection, and fixed-atom composition experiments in Section~\ref{app:search_diagnostics} use different historical task subsets and evaluation criteria. They are used only to localize failures to the search, candidate-coverage, and composition stages and are not merged with the four-seed main results. A low-budget pilot run using gplearn produced no tasks that passed the threshold, but its budget was substantially lower than the formal SRBench configuration; it is therefore not treated as evidence of the algorithm's capability.

\subsection{Retention of Standalone Successes by UNION}
\label{app:union_retention}

Table~\ref{tab:paired_counts} partitions the paired BANK and LLM outcomes but does not show whether UNION retains those successes. Table~\ref{tab:union_conversion} reports the conditional UNION success rate within each cell. If UNION simply selected the better standalone answer, the first three rows would be 100\% and the final row would be 0\%.

\begin{table*}[t]
\centering
\caption{Conditional UNION success rates (\%) within the four paired BANK/LLM outcome cells. Cell sizes are reported in Table~\ref{tab:paired_counts}.}
\label{tab:union_conversion}
{\small
\setlength{\tabcolsep}{3.5pt}
\begin{tabular}{lrrrrrrrr}
\toprule
&
\multicolumn{2}{c}{Acc\textsubscript{0.01}} &
\multicolumn{2}{c}{Acc\textsubscript{0.001}} &
\multicolumn{2}{c}{Strict ID+OOD} &
\multicolumn{2}{c}{SA} \\
\cmidrule(lr){2-3}\cmidrule(lr){4-5}\cmidrule(lr){6-7}\cmidrule(lr){8-9}
BANK/LLM outcome & L1 & L2 & L1 & L2 & L1 & L2 & L1 & L2 \\
\midrule
Both correct   & 100.0 & 100.0 & 98.6 & 99.0 & 98.5 & 100.0 & 100.0 & 100.0 \\
BANK only      & 94.6  & 98.4  & 95.1 & 99.6 & 92.8 & 100.0 & 90.7  & 99.0  \\
LLM only       & 0.0   & 0.0   & 20.0 & 20.0 & 33.3 & 0.0   & 0.0   & 0.0   \\
Both incorrect & 5.4   & 2.7   & 4.6  & 2.8  & 2.6  & 2.3   & 0.8   & 0.7   \\
\bottomrule
\end{tabular}
}

\end{table*}

UNION does not monotonically retain standalone successes. Under \(\operatorname{Acc}_{0.001}\), the L1 and L2 arms contain 20 and 15 LLM-only successes, but UNION retains only four and three; under SA, none are retained. UNION also loses some BANK successes and creates a few new successes when both standalone arms fail. The small full-library changes therefore reflect candidate competition, candidate interactions, and fitting-path changes rather than candidate availability alone.

\paragraph{Expression-Level Checks.}
BPG5 (seed 1) and PO10 (seed 2) are LLM-only successes that fail after UNION follows a different greedy path. Conversely, CRK3 (seed 1) and MatSci20 (seed 2) fail in both standalone arms but succeed after the combined pool supplies interacting decay or sine terms. For BPG10 (seed 0), UNION-Poly adopts the LLM-proposed \(P^a\) term and recovers \(P^{1/3}+P\). These examples illustrate candidate competition, candidate interaction, and a genuine library gap, respectively; they are not used to estimate effect sizes.

\subsection{Search, Selection, and Composition Diagnostics}
\label{app:search_diagnostics}

\paragraph{Stronger Combinatorial Search.}
We apply a higher-budget genetic-programming (GP) search to 30 tasks left unsolved by an earlier BANK run, while retaining generic mathematical primitives and adding no LLM candidates. Five GP seeds solve 19, 18, 16, 19, and 21 tasks, respectively (\(18.6\pm1.62\)).

\begin{table*}[t]
\centering
\caption{Paired counts for stronger GP and the current BANK method on 30 tasks left unsolved by an earlier BANK run.}
\label{tab:gp_bank_pairing}
{\small
\setlength{\tabcolsep}{5pt}
\begin{tabular}{lrrrrrr}
\toprule
Seed & BANK & GP & Both succeed & BANK only & GP only & Both fail \\
\midrule
0     & 1  & 19 & 0  & 1 & 19 & 10 \\
1     & 9  & 18 & 6  & 3 & 12 & 9  \\
2     & 8  & 16 & 6  & 2 & 10 & 12 \\
3     & 6  & 19 & 3  & 3 & 16 & 8  \\
Total & 24 & 72 & 15 & 9 & 57 & 39 \\
\bottomrule
\end{tabular}
}

\end{table*}

Across 120 task--seed pairs, GP uniquely recovers 57 pairs and BANK uniquely recovers nine. Thus, stronger semantics-free search can recover part of this preselected difficult subset. Because the subset was selected from an earlier failure set and the GP records do not retain expressions for strict ID+OOD or SA reevaluation, these counts cannot be merged into the benchmark-wide main score.

\paragraph{Fixed-Pool Selection.}
We hold each candidate pool fixed on 24 biological tasks and vary only the final selector. The comparison includes minimum training NMSE, POPPER-inspired counterexample falsification, process-reward reranking motivated by PRM methods, and a retrospective in-pool test upper bound. The upper bound asks whether any candidate in the pool passes the evaluation threshold; it is not a deployable method.

\begin{table}[t]
\centering
\caption{Selection rules evaluated on a fixed candidate pool.}
\label{tab:selection_methods}
{\small
\setlength{\tabcolsep}{5pt}
\begin{tabular}{lrr}
\toprule
Selection method & Successes & Tasks \\
\midrule
Minimum training NMSE & 10 & 24 \\
POPPER falsification  & 10 & 24 \\
PRM process reward    & 10 & 24 \\
In-pool test upper bound & 10 & 24 \\
\bottomrule
\end{tabular}
}

\end{table}

All four rules solve the same 10 tasks, so the remaining pools contain no candidate that passes the criterion. This one-pool diagnostic does not exclude a role for selection under noise or in denser candidate pools.

\paragraph{Fixed-Atom Composition.}
To distinguish missing atoms from failed composition, we hold the atomic vocabulary fixed and compare direct LLM composition with programmatic search.

\begin{table}[t]
\centering
\caption{Composition methods using the same atomic vocabulary.}
\label{tab:composition_methods}
{\small
\setlength{\tabcolsep}{2.5pt}
\begin{tabular}{lccc}
\toprule
Backbone & \shortstack{Direct LLM\\composition} &
\shortstack{Programmatic\\search} & Tasks \\
\midrule
Qwen3-Coder-480B & 0 & 5 & 10 \\
GPT-4.1          & 0 & 5 & 10 \\
GLM-4.6          & 0 & 7 & 10 \\
DeepSeek-V3.1    & 1 & 3 & 10 \\
\bottomrule
\end{tabular}
}

\end{table}

Programmatic search solves 3--7 tasks per record, compared with 0--1 for direct composition. The experiment contains only 10 tasks per backbone and no unified multi-seed repetition, so it localizes a possible composition bottleneck rather than ranking models.

\section{Extended Discussion and Limitations}
\label{app:extended_discussion}

\subsection{Implications for LLM-Based Symbolic Regression}

Our decomposition does not deny improvements achieved by existing LLM-based systems; it asks where each improvement occurs. DrSR constrains candidate generation using data-derived features; IGSR expands combinatorial search with Monte Carlo tree search and influence pruning; SR-LLM retrieves candidates from historical runs or external knowledge; and prior-guided methods use scientific constraints to suppress overfitting. Candidate generation, composition, final selection, and external-knowledge expansion are distinct capability sources and should not all be attributed to priors encoded in model parameters.

The published IGSR ablation illustrates the distinction. In its lung-cancer experiment, the complete method has test NMSE 0.000787; removing search yields 0.626, removing influence pruning yields 0.293, and removing both yields 4.85. Search-space expansion can therefore dominate reranking in that setting. Retrieval augmentation is another mechanism: retrieved operators or equation fragments can expand the candidate space at inference time. This can fill a coverage gap without implying that the same structure originated from an intrinsic model prior.

\subsection{Implications for Benchmark Design}

Benchmarks intended to measure search-boundary expansion should quantify semantics-free reachability before release. Generated tasks can first be stress-tested using fixed vocabularies and a prescribed search budget. Easily recovered tasks form a closed-vocabulary group; an open-vocabulary group can reserve operator families, named mechanisms, or higher-order interactions not disclosed to evaluated methods.

Future evaluations should report BANK, LLM, and UNION jointly and interpret paired gains together with task-level losses. Matched target and sham knockouts can distinguish removal of required structure from effects of candidate-set size, while OOD splits at multiple extrapolation distances can test interval sensitivity. This preserves resistance to complete-formula memorization while separating expression novelty from search-space novelty.

\subsection{Full Limitations}
\label{app:full_limitations}

The evidence has four limits. First, the matched knockout covers only 15 tasks and its \(\operatorname{Acc}_{0.01}\) confidence interval includes zero; it is directional evidence, whereas the polynomial control tests only a pronounced coverage gap. Second, although the BANK operators are publicly documented, the shallow expansion, interaction cap, greedy order, and fitting budget are our choices; the additional libraries reduce, but do not remove, this dependence. Third, the four-seed main results are stronger than the historical single-run and subset diagnostics, and GP uses a different budget and preselected tasks. Finally, strict recovery uses one OOD sample and threshold, while SA depends on expression parameterization, domain rules, and judge votes. The conclusions therefore apply only to the present LSR-Synth construction, vocabularies, and budgets; harder open-vocabulary settings may rely more strongly on scientific priors.


\end{document}